\documentclass[conference]{IEEEtran}
\ifCLASSINFOpdf
\else
\fi

\usepackage{graphicx}
\usepackage{url}
\usepackage{amsmath}
\usepackage{multirow}

\begin{document}
%
\title{An Integrated Framework for AI Assisted\\Level Design in 2D Platformers}

\author{
\IEEEauthorblockN{Antonio~Umberto~Aramini}
\IEEEauthorblockA{Politecnico di Milano, Milano, Italy\\
antonio.aramini@mail.polimi.it}
\and
\IEEEauthorblockN{Pier Luca Lanzi}
\IEEEauthorblockA{Politecnico di Milano, Milano, Italy\\
pierluca.lanzi@polimi.it}
\and
\and
\IEEEauthorblockN{Daniele Loiacono}
\IEEEauthorblockA{Politecnico di Milano, Milano, Italy\\
daniele.loiacono@polimi.it}
}


%


\raggedbottom

\maketitle

\begin{abstract}
The design of video game levels is a complex and critical task. Levels need to elicit fun and challenge while avoiding frustration at all costs.
In this paper, we present a framework to assist designers in the creation of levels for 2D platformers.
Our framework provides designers with a toolbox
	(i) to create 2D platformer levels,
	(ii) to estimate the difficulty and probability of success of single jump actions (the main mechanics of platformer games), and 
	(iii) a set of metrics to evaluate the difficulty and probability of completion of entire levels.
At the end, we present the results of a set of experiments we carried out with human players to validate the metrics included in our framework.
\end{abstract}


%
\IEEEpeerreviewmaketitle

\section{Introduction}
\label{sec:introduction}
Modern game development tools (like 
	\textit{Unreal Engine}\footnote{\url{https://www.unrealengine.com/}} and 
	\textit{Unity}\footnote{\url{https://unity3d.com/}})
	are powerful and flexible but constraint the actions of human designers
	to the positioning of objects (game elements) in scenes (levels) 
	and to the tweaking of parameters exposed by the developers (e.g. how fast an enemy runs or how much health it has).
These tools don't provide feedback neither about the functional characteristic of a level (e.g. its difficulty and feasibility), 
	nor about the type of player experience a level will enable.
Accordingly, designers either perform extensive playtest or build ad-hoc tools to support their content creation activities \cite{Skiddy}.

In the recent years, 
	artificial intelligence tools have been proposed to assist game designers in the creation of game content 
	like for example the generation of levels that affect players in terms of emotions 
	\cite{Platformer_Design_Session_Size_Seq_Features,Modeling_Experience_SMB} or 
	that satisfy structural constraints on the position of game elements
	\cite{Automatic_PCG_Platformers,Tanagra,Tanagra_Reactive_Planning_Constraints_Solver}.
In this context, level generation is usually based on design metrics extracted from the level structure (e.g., the position of game elements) or gameplay features (e.g., using data describing players' skill and playing style obtained with sessions of playtesting). 

In this paper, 
	we present a framework for the design of levels for 2D platformers that 
	extends previous approaches \cite{Platformer_Design_Session_Size_Seq_Features,Modeling_Experience_SMB} 
	by providing  immediate feedback about the \textit{functional} properties of levels such as 
	(i) the difficulty and probability of success of single jumps (the main mechanic of platformer games), and 
	(ii) a set of statistics to evaluate the difficulty and probability of completion of entire levels.

\newpage
Our framework has been developed as a modular extension of the popular Unity game engine and it is smoothly integrated in its editor.
It provides  a tool to design platformer levels using a set of structural features (e.g., fixed and moving platforms, deadly obstacles),
	to modify the available games mechanics (e.g., to specify whether the character can run, whether it can double jump or whether
	the players can control the character trajectory while jumping), and to modify the parameters of the underlying Unity 2D physics engine 
	(e.g. the gravity or the jump vertical takeoff speed).

The framework maps the structural features of a level into a graph in which 
	nodes represent platforms and 
	edges represent feasible jump trajectories between platforms.
Trajectories are computed in real time based on the structural properties of the level
	and on its functional gameplay properties (e.g., the character range of speed, the type of player controls, and the physics engine parameters).
Given two platforms (that is two nodes in the graph),
	an edge is created if there is a feasible jump that takes from one platform to the other one;
	each edge is characterized by a difficulty value and a probability of success.
The difficulty is computed by evaluating all the trajectories that start from an optimal position; 
	this is determined from the structure of the platforms
	and other functional properties such as the minimum and maximum speed that the character can reach in such position.
The probability of success is computed by evaluating a sample of trajectories starting from random points in the vicinity of the optimal position 
	generated using a noise function that models the player's skill level (the lower the noise the higher the skill).
Thus, the difficulty value takes into account mainly the structural properties of the level and the set up of the physics engine.
In contrast, the probability of success takes into account functional properties including the player skill.
The difficulty and the success probability are gameplay features obtained without actually performing sessions of playtesting 
	and provide key feedback on the designed content. 
These two metrics are combined to evaluate the difficulty and the probability of completion of entire levels. 
At the end, we performed a set of experiments involving human subject 
	to provide a preliminary validation of the our models of jump difficulty and success probability.
	
\newpage
Our approach extends previous works by considering a larger number of game elements (more platforms, obstacles, etc.) and 
	by intrinsically taking into account the parameters of the underlying physics engine so as to provide a more realistic model of the game mechanics.

\section{Related Work}
\label{sec:related}
Computer-aided design tools have become widely used in many production pipeline to reduce development time and costs.
These tools often involve the procedural generation of one or more types of artifact;
notable examples include \textit{SpeedTree}\footnote{\url{http://www.speedtree.com/}}, which allows the automatic generation of vegetation in 3D environments, \emph{Volcano}~\cite{Volcano}, that creates 3D models of swords by exploiting interactive evolution, and the \emph{Procedural Building Generator}~\cite{Designer_Driven_3D_Buildings}, that generates 3D buildings based on designer constraints and guidelines. 
In this section, we overview the most relevant work on difficulty metrics in video games and 
	on computer-aided design tools for content generation in 2D platformers;
we refer the reader to \cite{shaker2016procedural} for a more general and exhaustive discussion of procedural content generation (PCG) in games.

Smith et al. \cite{Framework_2D_Platformers} dissected level components in terms of their roles and structure to better understand the design behind levels and to provide designers a common vocabulary for the items forming the game worlds.
In a similar fashion, Dahlskog and Togelius \cite{Mario_W1L1} discussed the potential roles of design patterns in PCG, identifying them in levels of the original Super Mario Bros.; they also proposed how these patterns can be combined to create new levels that comply with specific constraints, while retaining variety of contents; this classification of patterns was then used to generate  levels by using micro-patterns as building blocks in a search-based algorithm~\cite{MultiLevel_Level_Generator}.
Pedersen et al.~\cite{Modeling_Experience_SMB} investigated the relationship between level design parameters, individual playing characteristics, and player experience; later extensions to this works included a much larger data set and an extensive validation through algorithms and human players \cite{Automatic_PCG_Platformers};
Nygren et al.~\cite{User_Preference_Level_Generation} investigated the design of levels with multiple paths and puzzle-based contents.
The approaches above have been further investigated in~\cite{Platformer_Design_Session_Size_Seq_Features} by analyzing how the session size, the different actions in game and the structural design patterns affect the players experience.
\emph{Launchpad}~\cite{Launchpad} and \emph{Tanagra}~\cite{Tanagra} applied the idea of \textit{rhythm group}: 
	a small level segment consisting of a rhythm of player actions (also referred as beats) and a level geometry that corresponds to that rhythm.
The Occupancy-Regulated Extension (ORE) algorithm introduced in~\cite{ORE}, relies on the idea of \emph{occupancy}, i.e., the positions (or \emph{anchors}) that player can occupy during gameplay; \emph{anchors} are thus used to iteratively expand the existing content by merging human-authored chunks.
Sorenson and Pasquier \cite{DBLP:conf/icccrea/SorensonP10} proposed a generative system for 2D platformer games in which creation process was driven by generic models of challenge-based fun based on existing theories of game design. They also presented a formulation of challenge of a jump between platforms $c(t)$, inspired to the work of Compton and Mateas \cite{DBLP:conf/aiide/ComptonM06}, that is proportional to the number of potential player trajectories that successfully traverse the gap. Mourato et al. \cite{Mourato2014Difficulty} extended the model of \cite{DBLP:conf/icccrea/SorensonP10} by adding the concept of time-based challenge, moving platforms, and opponents. They proposed a model of difficulty based on the probability of jump success as a function of (i) the maximum distance from the gap $d$ for which the jump is feasible and (ii) the player skill.
Ferreira et al. \cite{Multi_Pop_GA_Platformers} presented a multi-population genetic algorithm for generating levels by evolving four elements (terrain, enemies, coins and blocks), each one with its own encoding, population, and fitness function; at the end of the evolution the best four elements are combined to build the level.
Non-negative Matrix Factorisation (NMF) \cite{Combinatorial_PCG} follows a combinatorial approach and generates novel content through exploring new combinations of patterns using levels from five dissimilar content generators.
Reis et al. \cite{Human_Computation_PCG_Platformers} proposed a system that uses human computation to evaluate the quality of small portions of levels generated by another system and the tested sections are combined into a full level; their results show that the approach can create better levels both in terms of visual aesthetics and fun.
StreamLevels \cite{StreamLevels} supports the description of level structures using streamlines that 
	can be drawn or uploaded as real players trace data.
\textit{Super Mario Maker}, as a commercial game and design tool,  offers an easy and intuitive interface to create levels; each level is represented as a grid in which game elements (i.e. blocks, enemies, and power-ups), selected from a palette, can be placed. The design tools in the game do not include automated checks for playability or difficulty, thus the designer has to beat the levels in order to upload them on the Internet. For what concerns difficulty estimation, a clear rate (e.g. number of completions on number of tries) is displayed for each level and, once a sufficient number of players have played the level, an undisclosed algorithm, probably taking into consideration parameters such as the average number of deaths per player and their locations in the level, assigns a difficulty value from one (minimum difficulty, labeled as \textit{Easy}) to four (maximum difficulty, labeled as \textit{Very Hard}).


\section{Modeling 2D Platformers}
\label{sec:plaftormer}
Our framework captures the relevant features of platformer games by including the most common types of platforms, character movements, 
	and jump actions found in existing games. 

\subsection{Platforms}
Platforms are concrete blocks where the character can lean on, 
	and represent the fundamental elements that make it possible for the character to travel from one point to another in a level. 
In our framework, 
	platforms are defined by their \emph{position} and their \emph{length};
	they can be \textit{static}, having a fixed position, or they can be \textit{dynamic}, moving along a defined trajectory.
Platforms can also be \emph{fading} and as soon as the character reaches it, the platform starts to fade until it disappears letting the character fall;
they can have \emph{spikes} that hurt the character when it jumps on or move over them.
Finally, platforms can be tagged as \emph{starting} or \emph{exit} platform, if they are respectively the first or the last platform of the level.
All the platforms are characterized by a list of parameters that specify the platform type (e.g., fixed, moving, fading), 
	behavior (e.g., its moving trajectory, its moving or fading speed), and special properties 
	(e.g., whether it is a starting or ending platform, whether it is a checkpoint of some type).

\subsection{Player's Character Movement}
Jumping in platformer games is characterized by several design choices and 
	physics parameters \cite{Jump_Thesis}.
We selected a set of parameters and mechanics to build a framework that is as flexible as possible, while being also accessible to designers.
In particular, we modeled the movement of the character as an uniformly accelerated motion until either the maximum \emph{walking speed} or \emph{running speed} are reached.
The framework also allows to customize the acceleration of the character when a control input is given (i) to start the movement, (ii) to change the movement direction, and (iii) to stop the movement;
it is also possible to set  an \emph{infinite} acceleration, that is the character reaches the maximum speed instantly.
The character movement in the air is also modeled as
  an uniformly accelerated motion until the maximum \emph{air speed} is reached.
It is possible to customize the acceleration of the character
	while it is in the air, as previously described for the movements on the ground.
For jumps, we allow to customize the \emph{gravity constant}, the \emph{take off speed} (i.e., the initial vertical speed when the player's character jumps), and the maximum \emph{falling speed} (i.e., the maximum vertical speed of the player's character during a fall);
it is also possible to disable jump at all as well as to allow \emph{double jumps}, i.e., a second jump while the character is in the air after a previous jump.
Finally, it is possible to choose between \emph{static} or \emph{dynamic} jump models:
in the first case, the jump height is not influenced by how long the jump input is given (i.e., typically how long the jump button is pressed), so that every jump reaches the same height;
with a \emph{dynamic} jump model, if the jump button is released when the character is still moving upward, then his vertical speed is instantly set to zero, stopping his ascent;
this makes it possible for the player to control the height of each jump. 

\subsection{Level Navigation}
Given a level and a model of movement, 
	we can represent the navigation of the character through the level by
	computing a directed graph in which,
	nodes represent platforms and direct edges represent connections between platforms.
Note that, we use direct edges since a platform may be reachable from another platform, but not vice versa.

The level navigation directed graph is built as follows.
For each pair of platforms in the level, one tagged as \emph{starting} and one tagged as \emph{target}, 
	we check whether the character can reach the \emph{target} platform from the \emph{starting} platform;
for this purpose, 
	we generate a set of jump trajectories based on (i) all the movement parameters described above, 
		(ii) the position of the two platforms, and (iii) how the player approaches the jump, i.e., still, walking, running, or performing a double jump;
then, we check if the generated trajectories allows to reach the \emph{target} platform and, if such check is successful for at least one of the generated trajectories, we update the graph. We repeat this process for all the pairs of platforms.

\subsection{Jump Trajectories}
\begin{figure}[t]
	\centering
	\includegraphics[width=0.8\columnwidth,keepaspectratio]{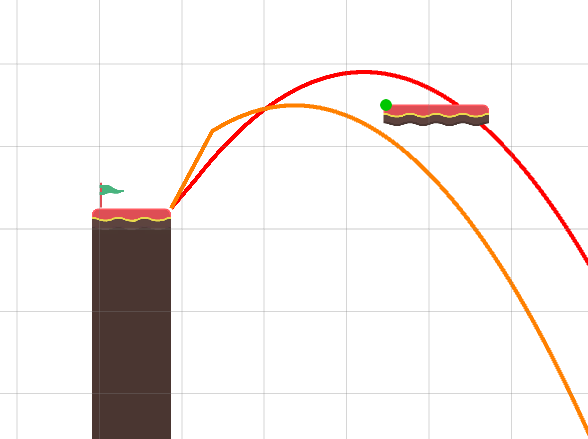}
	\caption{Two trajectories. Target point is shown in green. 
	The orange trajectory fails to reach the target point. 
	The red trajectory reaches the target point.}
	\label{fig:trajNonReachAndReach}
\end{figure}

A jump trajectory is defined by 
	(i)   the take off position, 
	(ii)  the jump direction,
	(iii) the take off speed vector,
	(iii) the state of the jump button (whether  it is pushed down).
Jump trajectories are represented as a list of piecewise defined functions $y_i(x)$ and the direction $d$:
	$(y_1(x),y_2(x),...,y_{n-1}(x),y_n(x),d)$.
Each segment identifies a different type of motion.
For example, the first segment $y_1(x)$ may correspond to the character moving 
	with a uniformly accelerated motion on the horizontal axis;
	while in the next segment $y_2(x)$, the character may be speed constant so that
	the functions describing the motion in the two intervals are different. 
Given a trajectory and the platform start position $x_{i,s}$, end position $x_{i,e}$ and ordinate $y_p$, 
	we say that a point $(x_p,y_p)$ is reachable if $y_i(x_p)>y_p$ and either $x_{i,s}\le x_p \le x_{i,e}$ 
	(if jumping to the right) or $x_{i,e}\le x_p \le x_{i,s}$ (if jumping to the left).
Basically, a point is reachable 
	if the character is in a more elevated position and horizontally aligned with the target. 
Figure \ref{fig:trajNonReachAndReach} shows two trajectories generated from the border 
	of the platform on the left and a target point (in green), 
	that is part of the border of the platform on the right; 
	the orange trajectory fails the reachability check, 
	whereas the red trajectory passes the reachability check as it is above the target.
	
\subsection{Generating Jump Trajectories}
\label{sec:trajectoryGeneration}
For each pair of \emph{starting} and \emph{target} platforms, 
	the framework generates a jump trajectory 
	based on the position and size of the two platforms.
We identified four types of jumps: 
	(i)    \emph{trivial}, 
	(ii)   \emph{simple}, 
	(iii)  \emph{falling}, and 
	(iv)   \emph{re-entrant}.	
\textit{Trivial jumps} identify the case of overlapping platforms (Figure~\ref{fig:trivialOptimalTraj}a)
	when it is almost impossible to fail: if the \emph{target} platform is above the \emph{starting} one,
	the character just needs to perform a jump while standing still and slightly 
	move toward the \emph{target} platform; 
	if the \emph{target} platform is below the \emph{starting} one, the character 
		just needs to fall straight to land on the \emph{target} (Figure~\ref{fig:trivialOptimalTraj}a).
%

\begin{figure}[t]
	\centering
	\includegraphics[width=0.8\columnwidth,keepaspectratio]{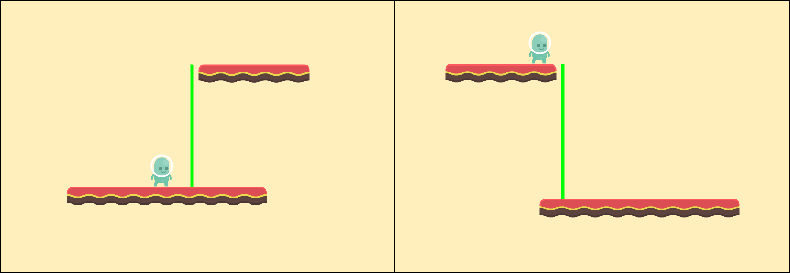}
	
	\centerline{(a)}
	~\\
	
	\centering
	\includegraphics[width=0.8\columnwidth,keepaspectratio]{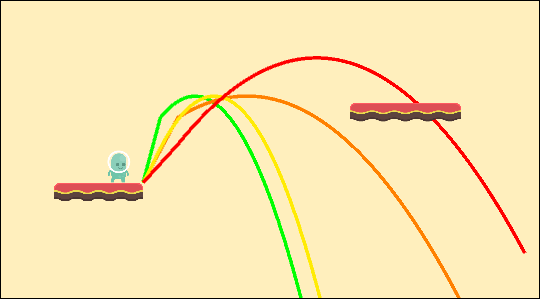}

	\centerline{(b)}
	~\\
	
	\includegraphics[width=0.8\columnwidth,keepaspectratio]{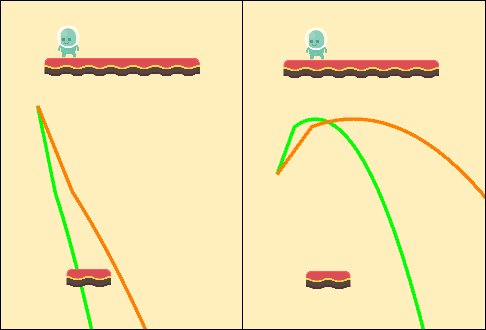}
	
	\centerline{(c)}
	~\\
	
	\includegraphics[width=0.8\columnwidth,keepaspectratio]{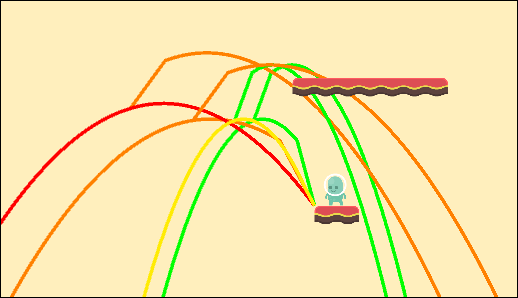}
	
	\centerline{(d)}
	~\\
	
	\caption{Generated trajectories for (a) trivial, (b) simple, (c) falling, and (d) re-entrant jumps.
	Jump trajectories are green when the character is still;
		they are yellow, when the takeoff speed is positive but the character is not running;
		they are orange, when the takeoff speed is zero and the character is running;
		they are red, when the takeoff speed is positive and the character is running.}
%
	
	\label{fig:trivialOptimalTraj}
	\label{fig:genericTrajSingleJump}
	\label{fig:reentrantTraj1}
	\label{fig:fallingTrajSingleAndDoubleJump}
	\label{fig:genericTrajDoubleJump}
	\label{fig:trivialOptimalTraj}
\end{figure}

\textit{Simple jumps} identify generic jump configurations when the platforms do not overlap 
	(Figure~\ref{fig:genericTrajSingleJump}b).
To generate the jump trajectories, the takeoff point is selected among the 
	\emph{left} and the \emph{right} edges of the \emph{starting} platform,
	based on the position of the \emph{target} platform.
The vertical takeoff speed depends on 
	whether (i) the character approaches the jump standing still (the horizontal speed is zero)
	     or (ii) the character is moving to the takeoff point;
	whether (iii) the running input button is pushed (on) or (iv) not (off).
Accordingly, four jump trajectories are generated (Figure~\ref{fig:genericTrajSingleJump}b) 
%
%
If \emph{double jump} is enabled, for each one of the four trajectories, 
	two additional trajectories are generated by assuming that the second jump is performed either
	at the apex of the first jump or when the character reaches the same vertical coordinate 
	of the initial takeoff point.
Overall, up to twelve trajectories can be generated for a \emph{simple} jump: four for a single jump and eight for double jumps. 
%

When the \emph{target} platform is below the \emph{starting} platform and the latter 
	covers the former completely, the character performs a \textit{falling jump} (Figure~\ref{fig:fallingTrajSingleAndDoubleJump}c).
%
%
In this case, the takeoff point is selected among the edges of the \emph{starting} platform, 
	choosing the closest one to the \emph{target} platform.
Two jump trajectories are generated assuming that the characters simply 
	falls down from the takeoff point with a takeoff speed equal to zero.
When \emph{double jump} is enabled, 
	two additional jump trajectories are generated if \emph{double jump} is enabled as follows.
%

When the \emph{target} platform is above the \emph{starting} platform and
	the horizontal projection of the former covers the latter (Figure~\ref{fig:reentrantTraj1}d), 
	the character will perform a \textit{re-entrant jump}: 
	first it has to jump away from the \emph{starting} platform;
	next it jumps back toward the \emph{target} platform.
Re-entrant jumps are possible only if \emph{double jump} is enabled;
	the takeoff point is selected among the edges of the \emph{starting} platform, 
	by choosing the closest one to the \emph{target} platform.
Jump trajectories are generated similarly to the case of \emph{simple} jumps
	by assuming that the character jumps standing still or while running toward the takeoff point.
In the latter case, we compute the horizontal takeoff speed, as the highest speed 
	that the character can reach.
Figure~\ref{fig:reentrantTraj1}d shows an example of the four jump trajectories generated.
%

The approach we described assumes that the platforms are static. When dynamic platforms are involved, 
	the bounding boxes of the moving platforms are
	discretized according to the direction of their trajectories.
Then, the reachability analysis is performed considering all the possible positions of the 
	dynamic platforms within its bounding box using the platform width or height as a step
  between adjacent positions.
%

\section{Modeling Jump Success Probability}
\label{sec:modeling}
Players can perform the same jump between two platforms in many different ways,
	thus the same jump can follow several different trajectories.
In our framework, the success probability of a jump is evaluated by applying a noise function to generate, 
	for each jump trajectory, a set of random takeoff positions around the optimal takeoff position. 
In particular, our framework supports three noise functions:
	\textit{Uniform}, \textit{Gaussian without resampling} and \textit{Gaussian with resampling}.
The success probability of the jump is estimated as the percentage of successful sampled trajectories. 
The noise functions have several parameters such as
	(i)   the average reaction time of the player \textit{Rt} (ranging between 0.01s and 1s);
	(ii)  the player skill value \textit{Ps} (with values from 1 to 50); 
	(iii) the absolute value of either the horizontal speed $\lvert v_x \rvert$ or the vertical speed $\lvert v_y \rvert$
	      at which the character approaches the jump.
These parameters can be set by the designer or empirically estimated.
Intuitively, the higher \textit{Rt} and $\lvert v_x \rvert$ (or $\lvert v_y \rvert$) are,
	the more the noise affects the sample trajectories, moving them away from the optimal ones;
	the higher \textit{Ps} is, the lower the effect of the noise becomes.

For each random takeoff point, the framework generates a sample of trajectories;
	a jump difficulty coefficient takes 
		into account the level of challenge 
		that different types of the two platforms might introduce
		(e.g., two moving and distant platforms compared to two static platforms);
	the success probability 
		is then computed as the percentage of successful sampled trajectories
		weighted by the inverse of the jump difficulty coefficient.
Note that, if the optimal jump trajectory is unsuccessful (i.e., there are no feasible jumps), 
	the framework counts the respective sample trajectories as unsuccessful without actually generating them. 
Furthermore, if the target platform is \textit{dynamic} and moves along a trajectory, 
	the success probability is computed as the average of the estimated probabilities obtained 
	by using the target optimal position (i.e., the position resulting in the
	lowest difficulty) and the adjacent ones.
In the following, we present and example of how success probability is computed for simple jumps; we refer the reader to \cite{aramini:2017:thesis}
	for an extended discussion of how probabilities are computed for all the possible jump and platform types combinations.

\subsection{Evaluating Simple Trajectory Jumps}
In simple jumps, the noise functions affect the horizontal 
	coordinate of the optimal takeoff point.
When using the uniform noise model, 
	the random takeoff points are selected with a uniform probability distribution from an interval 
	of size $\delta$ around the optimal position. 
When using the \textit{Gaussian without resampling} or the \textit{Gaussian with resampling} noise functions,
	the random takeoff points are sampled from a Gaussian distribution.
The two Gaussian noise functions differ in how they manage the sampling of takeoff 
	positions near the platform border which might actually be outside the platform. 
The \textit{Gaussian without resampling} noise function counts the takeoff points that are off the starting platform as unsuccessful sample trajectories, whereas the \textit{Gaussian with resampling} noise function generates a new random takeoff point until it is on the starting platform. 
Thus, the former models the possibility of failing a jump because the player pressed the jump button too late while moving toward the optimal takeoff point. 

\begin{figure}
	\centering

	\includegraphics[width=0.8\columnwidth,keepaspectratio]{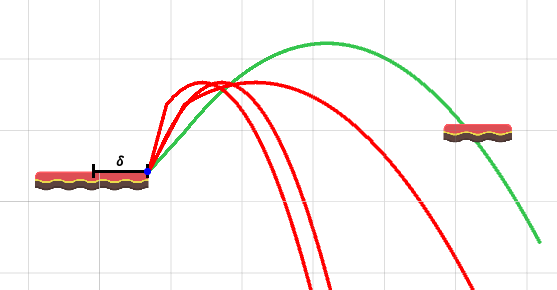}
	
	\centerline{(a)}
	~\\
	
	\includegraphics[width=0.8\columnwidth,keepaspectratio]{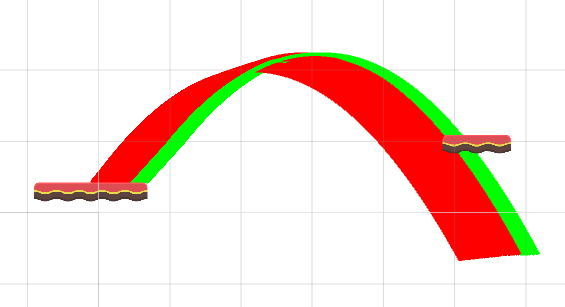}

	\centerline{(b)}
	~\\

	\includegraphics[width=0.8\columnwidth,keepaspectratio]{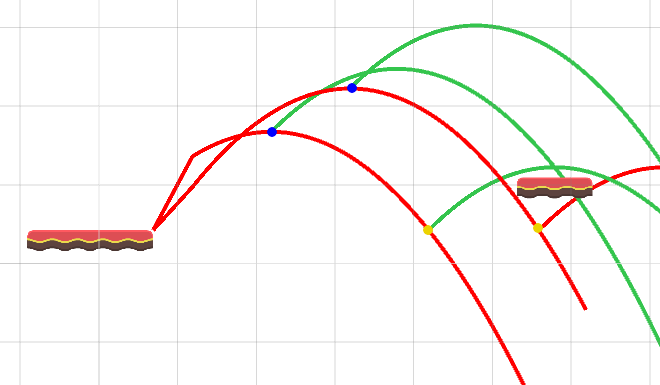}

	\centerline{(c)}
	~\\

	\includegraphics[width=0.8\columnwidth,keepaspectratio]{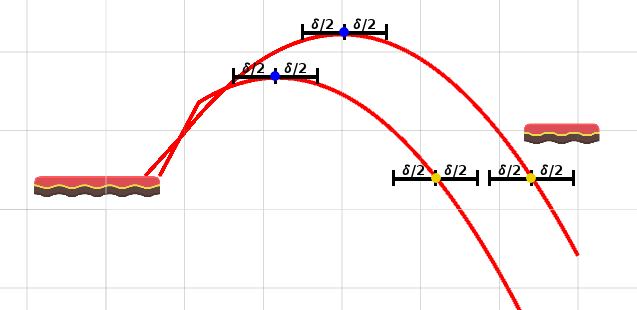}
	
	\centerline{(d)}
	~\\
	
	\includegraphics[width=0.8\columnwidth,keepaspectratio]{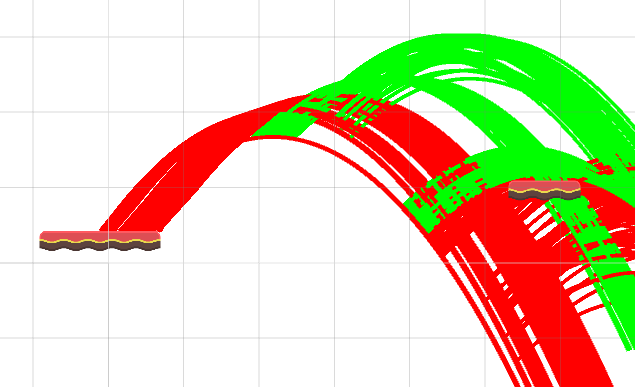}

	\centerline{(e)}
	~\\

	\caption{(a) Boundary trajectories for a simple jump configuration;
		the only successful boundary trajectory is the one in green;
		the optimal takeoff point is highlighted in blue. 
		(b) The generated sample trajectories for the scenario using a \textit{Uniform} noise function.}
	\label{fig:uniformFirstJumpInterval}
	\label{fig:uniformSamplesFirstJump}
	\label{fig:simpleDoubleJumpRUNOptimals}
	\label{fig:uniformSimpleDoubleJumpSampleIntervals}
	\label{fig:uniformSimpleDoubleJumpSamples}
\end{figure}

%

Figure \ref{fig:uniformFirstJumpInterval}a reports the four trajectories for a simple jump scenario
	defined using different parameters (Section \ref{sec:trajectoryGeneration}).
The red trajectories fail to reach the target platform, whereas the green one is successful;
	the optimal takeoff point is highlighted with a blue dot.
The interval $\delta$ associated to the successful optimal trajectory is displayed in black. Figure \ref{fig:uniformSamplesFirstJump}b shows the trajectories sampled using the \textit{Uniform} noise function---in red the unsuccessful ones, in green the successful ones. 
The estimated probability of success for the jump is 0.075. 

If double jump is enabled, the character can perform a second jump while in the air and 
	the framework will take into account the possibility of this event.
Figure \ref{fig:simpleDoubleJumpRUNOptimals}c shows, for the sake of simplicity, only the optimal trajectories that involve the player pressing the run button; for each initial jump trajectory, we show the corresponding two double jump trajectories: one starting at the peak of the first trajectory (the blue dot),
	for reaching the maximum height possible; 
	one starting at the same $y$ coordinate of the first takeoff (the yellow dot),
	for reaching the longest distance possible.

Figure \ref{fig:uniformSimpleDoubleJumpSampleIntervals}d shows two trajectories for the initial jump and highlights
	the sampling intervals for the takeoff position of the second jump. 
Note that, we sample the $x$ coordinate of the double jump random takeoff points from these intervals, whereas the $y$ coordinate is obtained evaluating the respective first jump trajectory in the sampled $x$ coordinate.
Figure \ref{fig:uniformSimpleDoubleJumpSamples}e shows the sample trajectories obtained using the \textit{Uniform} noise function to generate the random takeoff points; the estimated probability of success for the jump in this scenario is 0.208, which is not affected by the difficulty coefficient.
\section{Experimental Validation}
\label{sec:experiments}
We performed a set of experiments to validate our model for the estimation of the success probability of jumps and to select
	the most adequate noise model. 
We used our tool to build a simple game and deployed it online as a web app. The game was structured in 16 \textit{screens}, each one asking the players to jump across two platforms organized in different ways: (i) 2 screens involved \textit{trivial} jumps across two nearby platforms with no possibility of failing; (ii) 6 screens involved \textit{simple} trajectory jumps across two platforms, which were positioned to have a gap of variable width between them; (iii) 4 screens involved \textit{Falling} trajectories with the starting platform positioned above the target one; (iv) 4 screens involved \textit{Reentrant} jumps across two platforms, with the starting platform below the target one, requiring good skills and timing. The player could walk, run, jump or double jump. 

We asked users with different skill levels to perform from 10 up to 50 jumps for each session. Users could repeat the experiment as many times as they wanted. At first, users were asked to complete a sequence of warm up jumps, randomly selected from the 16 screens; the same jump was never presented twice in a row. Next, users were presented with the actual series of jumps and at the end they had the opportunity to fill an optional form to provide some information about their age, how much they liked platformers and how difficult the jump trials felt. For each jump we recorded (i) the identifier of the screen (\textit{screen ID}); (ii) the coordinates of the takeoff position; (iii) the coordinates of the landing position (only if the user managed to land on the target platform); (iv) the horizontal speed at which the character took off; (v) whether the target platform was successfully reached; and (vi) the jump trajectory.

The experiment involved 58 users; 26 of them also completed the form; the users who provided additional information were between 20 and 40 years old and reported an above the average liking of platformers. Overall, we recorded information about 2361 jumps, 1477 of which performed by the 26 users who also completed the form.

Table \ref{tab:allDataProbabilities} reports a summary of the information collected with the experiments for each jump screen (identified by \textit{screen ID}) available to the players: the jump type (\textit{Trajectory Type}), the total number of times the jump screen was presented to the users (\textit{\#Jumps}), the total number of times the jump screen was successfully completed by the users (\textit{\#Successes}) and the ratio of successes for the jump screen (\textit{\%Success}). All jump trials reported an experimental success probability greater than 30\% except trials \#4 and \#8 that were completed 21.6\% of the times. The former required jumping while running and performing a well timed double jump; the latter was a \textit{Reentrant} jump involving featuring a moving starting platform.
The two trivial jumps (\#9 and \#12) reported a success probability close to one; in particular, jump \#12 was failed only once out of 157 tries.
Re-entrant configurations appears to be the hardest to tackle, whereas falling configurations appeared to be the easiest ones.

\begin{table}
\caption{A summary of the data collected for each jump trial.}
\label{tab:allDataProbabilities}
\centering
\begin{tabular}{|c|c|c|c|c|}
	\hline
	\textbf{screen ID}  & \textbf{Trajectory Type} & \textbf{\#Jumps} & \textbf{\#Successes} & \textbf{\%Success} \\	\hline
	0  & Simple        & 156              & 136               & 0.872 \\	\hline
	1  & Simple        & 131              & 80                & 0.611 \\	\hline
	2  & Simple        & 158              & 114               & 0.722 \\	\hline
	3  & Simple        & 159              & 104               & 0.654 \\	\hline
	4  & Simple        & 153              & 33                & 0.216 \\	\hline
	5  & Reentrant      & 150              & 53                & 0.353 \\	\hline
	6  & Reentrant      & 139              & 76                & 0.547 \\	\hline
	7  & Reentrant      & 134              & 43                & 0.321 \\	\hline
	8  & Reentrant      & 148              & 32                & 0.216 \\	\hline
	9  & Trivial        & 157              & 139               & 0.885 \\	\hline
	10 & Falling        & 151              & 142               & 0.940 \\	\hline
	11 & Falling        & 156              & 143               & 0.917 \\	\hline
	12 & Trivial        & 157              & 156               & 0.994 \\	\hline
	13 & Falling        & 140              & 128               & 0.914 \\	\hline
	14 & Falling        & 141              & 93                & 0.660 \\	\hline
	15 & Simple        & 131              & 64                & 0.489 \\	\hline
\end{tabular}
\end{table}

We estimated the success probability of each configuration using the three noise models and employing different combinations of average reaction time \textit{Rt} and player skill \textit{Ps}. We compared the probabilities estimated by our framework against the experimental data collected using the \textit{Mean Absolute Error} (MAE).
%

\begin{figure}
	\centering
		\includegraphics[width=0.3\textwidth,keepaspectratio]{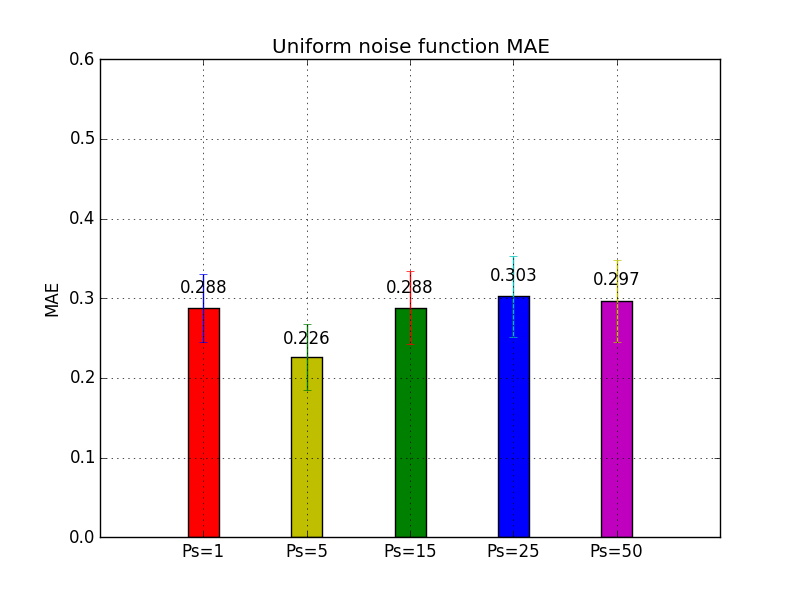}
		
		\centerline{(a)}
		~\\
		
		\includegraphics[width=0.3\textwidth,keepaspectratio]{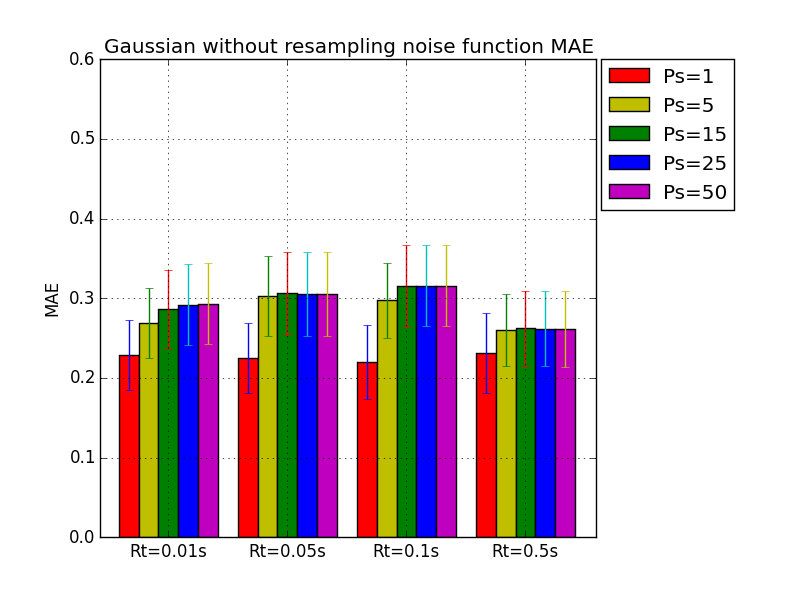}
		
		\centerline{(b)}
		~\\

		\includegraphics[width=0.3\textwidth,keepaspectratio]{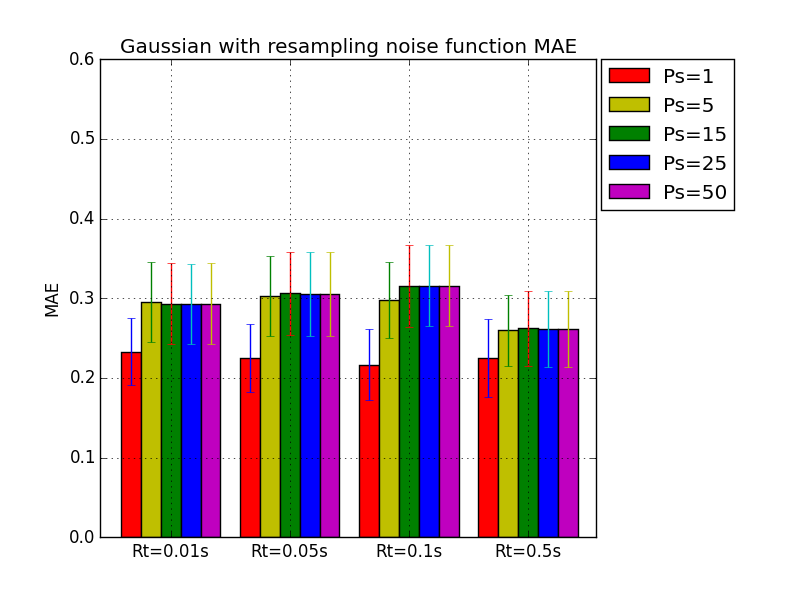}
		
		\centerline{(c)}
		~\\
	
	\caption{Bar chart comparing Mean Absolute Error (MAE) values for  (a) the \textit{Uniform} noise function; 
		(b) the \textit{Gaussian without resampling} noise function;
		(c) the \textit{Gaussian with resampling} noise function.}
	\label{fig:uniformMAEChart}
	\label{fig:gaussianNoResMAEChart}
	\label{fig:gaussianResMAEChart}
\end{figure}

Figure \ref{fig:uniformMAEChart}a shows MAE values for the \textit{Uniform} noise function.
Tables \ref{tab:gaussianModelsMAE1} and \ref{tab:gaussianModelsMAE2} report MAE values  and the respective standard error computed for the two Gaussian noise models according to different combinations of \textit{Rt} and \textit{Ps}. Figures \ref{fig:gaussianResMAEChart}b and \ref{fig:gaussianNoResMAEChart}c compare MAE values for each Gaussian noise model.


By observing the performance metrics, we see that the three noise models keep MAE values below 0.32. In particular, for the \textit{Uniform} noise model, the minimum is 0.227, corresponding to $Ps=5$; for what concerns the Gaussian models, the minimum MAE value is 0.217, corresponding to the \textit{Gaussian with resampling} noise function with $Ps=1$ and $Rt=0.1s$.


\begin{table}
	\caption{Mean Absolute Error (MAE) values for the \textit{Gaussian without resampling} (GNR) and the \textit{Gaussian with resampling} (GR) noise functions with respect to different combinations of $Rt$ and $Ps$.}
	\tiny
	\centering
	\begin{tabular}{|l|c|c|c|c|}
		\hline
		\multirow{2}{*}{\textbf{MAE}} & \multicolumn{2}{c|}{Rt=0.01s} & \multicolumn{2}{c|}{Rt=0.05s} \\ \cline{2-5}
		& \textbf{GNR} & \textbf{GR}    & \textbf{GNR} & \textbf{GR}    \\ \hline
		Ps=1                    & $0.229 \pm 0.044$ & $0.233 \pm 0.042$ & $0.225 \pm 0.044$ & $0.224 \pm 0.043$  \\ \hline
		Ps=5                    & $0.269 \pm 0.044$ & $0.295 \pm 0.050$ & $0.302 \pm 0.050$ & $0.303 \pm 0.051$  \\ \hline
		Ps=15                    & $0.286 \pm 0.048$ & $0.293 \pm 0.051$ & $0.306	\pm 0.053$ & $0.306 \pm	0.053$  \\ \hline
		Ps=25                    & $0.292 \pm 0.050$ & $0.293 \pm 0.051$ & $0.306	\pm 0.053$ & $0.306 \pm	0.053$  \\ \hline
		Ps=50                    & $0.293 \pm 0.051$ & $0.293 \pm 0.051$ & $0.306	\pm 0.053$ & $0.306	 \pm 0.053$  \\ \hline
	\end{tabular}
	\label{tab:gaussianModelsMAE1}
\end{table}

\begin{table}
	\caption{Mean Absolute Error (MAE) values for the \textit{Gaussian without resampling} (GNR) and the \textit{Gaussian with resampling} (GR) noise functions with respect to different combinations of $Rt$ and $Ps$.}

	\tiny
	\centering
	\begin{tabular}{|l|c|c|c|c|}
		\hline
		\multirow{2}{*}{\textbf{MAE}} & \multicolumn{2}{c|}{Rt=0.1s} & \multicolumn{2}{c|}{Rt=0.5s} \\ \cline{2-5}
		& \textbf{GNR} & \textbf{GR}    & \textbf{GNR} & \textbf{GR}    \\ \hline
		Ps=1                    & $0.220 \pm 0.046$ & $0.217 \pm 0.045$ & $0.231 \pm 0.050$ & $0.225 \pm 0.049$  \\ \hline
		Ps=5                    & $0.297 \pm 0.047$ & $0.298 \pm 0.047$ & $0.260 \pm 0.045$ & $0.260 \pm 0.045$  \\ \hline
		Ps=15                    & $0.315 \pm 0.051$ & $0.315 \pm 0.051$ & $0.262 \pm 0.047$ & $0.262 \pm 0.047$  \\ \hline
		Ps=25                    & $0.316 \pm 0.051$ & $0.316 \pm 0.051$ & $0.262 \pm 0.047$ & $0.262 \pm 0.047$  \\ \hline
		Ps=50                    & $0.316 \pm 0.051$ & $0.316 \pm 0.051$ & $0.261 \pm 0.047$ & $0.261 \pm 0.047$  \\ \hline
	\end{tabular}
	\label{tab:gaussianModelsMAE2}
\end{table}

\subsection{Level difficulty and probability of success}\label{sec:levelsDiffAndProb}
We extended the evaluation of difficulty and success probability from single jumps to the case of entire levels 
	viewed as multiple sequences of consecutive jumps.
We defined a \textit{path} as the sequence of jumps that the character performs to move through a level; 
	a path is list of directed edges $(e_1,e_2,...,e_L-1,e_L)$.
Note that we only consider acyclic paths and thus a node of the graph (a platform) can be visited only once in a path.
We also assume that each jump is independent from the others, accordingly we evaluate success probability and difficulty of a path $P$ as, 
\begin{equation}
	p(P)=\prod_{e_i \in P}p(e_i)
	\label{eq:pathProbability}
\end{equation}
\begin{equation}
	d_{t}(P)=\sum_{e_i \in P}d(e_i)
	\label{eq:pathTotalDifficulty}
\end{equation}
The probability of success $p$ of a path $P$ (equation \ref{eq:pathProbability}) is computed as the product of the success probabilities of the jumps in the path (all the edges $p(e_i)$). The difficulty $d_t$ of a path $P$ (equation \ref{eq:pathTotalDifficulty}) is the sum of the edge difficulties $d(e_i)$. The $d_t$ metric is agnostic to the position of the jump in the level (e.g. the first jump has the same weight of the last jump in the path), thus it provides a rough estimate of the path difficulty also in terms of its length; $d_t$ can also be viewed as a metric to roughly estimate the player's fatigue.

A level can be traversed in multiple ways from start to finish, accordingly, our framework applies depth first search on the graph representing the level to extract the paths that are feasible without dying (e.g., because failing a jump or by exhausting the character health because of harmful obstacles like spikes).
Our framework supports the visualization of the minimum difficulty path, where the difficulty metric can be selected by the designer among the set of metrics that we previously discussed.
\begin{figure}[t]
	\centering
	\includegraphics[width=\columnwidth,keepaspectratio]{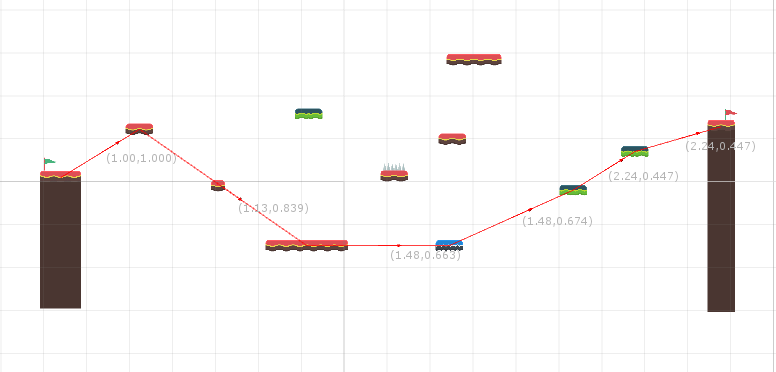}
	\caption{An example of minimum difficulty path (in red) displayed by the framework.}
	\label{fig:minDiffPathExample}
\end{figure}
Figure \ref{fig:minDiffPathExample} shows the minimum difficulty path (in red) with respect to the $d_t$ metric for an example level; each one of the edge composing the path is displayed together with its difficulty and probability of success.
%
%
\section{Conclusions}
\label{sec:conclusions}
We presented a framework for the design and evaluation of levels for 2D platformers
	that we developed as a modular extension of a popular game engine Unity.
The framework takes into account the \textit{structural features} of levels (e.g., the platforms' types and positions) 
	and the \textit{functional features} of the level (e.g., the underlying physics, the character moving and jumping capabilities)
	to evaluate levels in real time in terms of difficulty and probability of success for single jumps and for the whole level.
We performed a preliminary set of experiments involving human players to validate the approach we use to evaluate the difficulty and the probability of success of single jump actions. 

We plan to extend this work in two ways.
First, we want to validate the method proposed for the evaluation of entire levels by collecting gameplay data with human players using levels with different characteristics.
Next, we plan to add a search-based procedural content generator in order to let designers create levels with specific properties or metrics values.

\begin{thebibliography}{10}
\providecommand{\url}[1]{#1}
\csname url@samestyle\endcsname
\providecommand{\newblock}{\relax}
\providecommand{\bibinfo}[2]{#2}
\providecommand{\BIBentrySTDinterwordspacing}{\spaceskip=0pt\relax}
\providecommand{\BIBentryALTinterwordstretchfactor}{4}
\providecommand{\BIBentryALTinterwordspacing}{\spaceskip=\fontdimen2\font plus
\BIBentryALTinterwordstretchfactor\fontdimen3\font minus
  \fontdimen4\font\relax}
\providecommand{\BIBforeignlanguage}[2]{{%
\expandafter\ifx\csname l@#1\endcsname\relax
\typeout{** WARNING: IEEEtran.bst: No hyphenation pattern has been}%
\typeout{** loaded for the language `#1'. Using the pattern for}%
\typeout{** the default language instead.}%
\else
\language=\csname l@#1\endcsname
\fi
#2}}
\providecommand{\BIBdecl}{\relax}
\BIBdecl

\bibitem{Skiddy}
\BIBentryALTinterwordspacing
A.~Baldiraghi, ``Simple mechanics, complex puzzle creation. can computer-aided
  puzzle design help us create fair, fun and interesting challenges?''
  [Online]. Available:
  \url{https://www.gamasutra.com/blogs/AndreaBaldiraghi/20160810/278907/}
\BIBentrySTDinterwordspacing

\bibitem{Platformer_Design_Session_Size_Seq_Features}
N.~Shaker, G.~N. Yannakakis, and J.~Togelius, ``Digging deeper into platform
  game level design: Session size and sequential features,'' \emph{Proceedings
  of the European Conference on Applications of Evolutionary Computation
  (EvoApplications)}, 2012.

\bibitem{Modeling_Experience_SMB}
C.~Pedersen, J.~Togelius, and G.~N. Yannakakis, ``Modeling player experience in
  super mario bros,'' \emph{IEEE Conference on Computational Intelligence and
  Games}, 2009.

\bibitem{Automatic_PCG_Platformers}
N.~Shaker, G.~N. Yannakakis, and J.~Togelius, ``Towards automatic personalized
  content generation for platform games,'' \emph{Proceedings of the Sixth AAAI
  Conference on Artificial Intelligence and Interactive Digital Entertainment},
  2010.

\bibitem{Tanagra}
G.~Smith, J.~Whitehead, and M.~Mateas, ``Tanagra: A mixed-initiative level
  design tool,'' \emph{IEEE Transactions on Computational Intelligence and AI
  in Games}, 2010.

\bibitem{Tanagra_Reactive_Planning_Constraints_Solver}
------, ``Tanagra: Reactive planning and constraint solving for
  mixed-initiative level design,'' \emph{IEEE Transactions on Computational
  Intelligence and AI in Games}, 2011.

\bibitem{Volcano}
D.~Loiacono, R.~Mainetti, and M.~Pirovano, ``Volcano: An interactive sword
  generator,'' \emph{Games, Entertainment, and Media}, 2015.

\bibitem{Designer_Driven_3D_Buildings}
J.~M. Pe{\~n}a, J.~Viedma, S.~Muelas, and A.~L.~L. Pe{\~n}a, ``Designer-driven
  3d buildings generated using variable neighborhood search,'' \emph{IEEE
  Conference on Computational Intelligence and Games}, 2014.

\bibitem{shaker2016procedural}
N.~Shaker, J.~Togelius, and M.~J. Nelson, \emph{Procedural Content Generation
  in Games: A Textbook and an Overview of Current Research}.\hskip 1em plus
  0.5em minus 0.4em\relax Springer, 2016.

\bibitem{Framework_2D_Platformers}
G.~Smith, M.~Cha, and J.~Whitehead, ``A framework for analysis of 2d platformer
  levels,'' \emph{Proceedings of the 2008 ACM SIGGRAPH symposium on Video
  games}, 2008.

\bibitem{Mario_W1L1}
S.~Dahlskog and J.~Togelius, ``Patterns and procedural content generation
  revisiting mario in world 1 level 1,'' \emph{Proceedings of the Workshop on
  Design Patterns in Games (DPG 2012)}, 2012.

\bibitem{MultiLevel_Level_Generator}
------, ``A multi-level level generator,'' \emph{IEEE Conference on
  Computational Intelligence and Games}, 2014.

\bibitem{User_Preference_Level_Generation}
N.~Nygren, J.~Denzinger, B.~Stephenson, and J.~Aycock, ``User-preference-based
  automated level generation for platform games,'' \emph{IEEE Conference on
  Computational Intelligence and Games}, 2011.

\bibitem{Launchpad}
G.~Smith, J.~Whitehead, M.~Mateas, M.~Treanor, J.~March, and M.~Cha,
  ``Launchpad: A rhythm-based level generator for 2-d platformers,'' \emph{IEEE
  Transactions on Computational Intelligence and AI in Games}, 2011.

\bibitem{ORE}
P.~Mawhorter and M.~Mateas, ``Procedural level generation using
  occupancy-regulated extension,'' \emph{IEEE Conference on Computational
  Intelligence and Games}, 2010.

\bibitem{DBLP:conf/icccrea/SorensonP10}
\BIBentryALTinterwordspacing
N.~Sorenson and P.~Pasquier, ``The evolution of fun: Automatic level design
  through challenge modeling,'' in \emph{Proceedings of the International
  Conference on Computational Creativity, Lisbon, Portugal, January 7-9,
  2010.}, D.~Ventura, A.~Pease, R.~P. y~P{\'{e}}rez, G.~Ritchie, and T.~Veale,
  Eds.\hskip 1em plus 0.5em minus 0.4em\relax computationalcreativity.net,
  2010, pp. 258--267. [Online]. Available:
  \url{http://computationalcreativity.net/iccc2010/papers/sorenson-pasquier.pdf}
\BIBentrySTDinterwordspacing

\bibitem{DBLP:conf/aiide/ComptonM06}
\BIBentryALTinterwordspacing
K.~Compton and M.~Mateas, ``Procedural level design for platform games,'' in
  \emph{Proceedings of the Second Artificial Intelligence and Interactive
  Digital Entertainment Conference, June 20-23, 2006, Marina del Rey,
  California}, J.~E. Laird and J.~Schaeffer, Eds.\hskip 1em plus 0.5em minus
  0.4em\relax The {AAAI} Press, 2006, pp. 109--111. [Online]. Available:
  \url{http://www.aaai.org/Library/AIIDE/2006/aiide06-022.php}
\BIBentrySTDinterwordspacing

\bibitem{Mourato2014Difficulty}
\BIBentryALTinterwordspacing
F.~Mourato, F.~Birra, and M.~P. dos Santos, ``Difficulty in action based
  challenges: Success prediction, players' strategies and profiling,'' in
  \emph{Proceedings of the 11th Conference on Advances in Computer
  Entertainment Technology}, ser. ACE '14.\hskip 1em plus 0.5em minus
  0.4em\relax New York, NY, USA: ACM, 2014, pp. 9:1--9:10. [Online]. Available:
  \url{http://doi.acm.org/10.1145/2663806.2663832}
\BIBentrySTDinterwordspacing

\bibitem{Multi_Pop_GA_Platformers}
L.~Ferreira, L.~Pereira, and C.~Toledo, ``A multi-population genetic algorithm
  for procedural generation of levels for platform games,'' \emph{Proceedings
  of the 2014 Conference Companion on Genetic and Evolutionary Computation
  Companion (GECCO Comp '14)}, 2014.

\bibitem{Combinatorial_PCG}
N.~Shaker and M.~Abou-Zleikha, ``Alonewe can do so little, togetherwe can do so
  much: A combinatorial approach for generating game content,''
  \emph{Conference on Artificial Intelligence and Interactive Digital
  Entertainment}, 2014.

\bibitem{Human_Computation_PCG_Platformers}
W.~M.~P. Reis, L.~H.~S. Lelis, and Y.~Gal, ``Human computation for procedural
  content generation in platform games,'' \emph{IEEE Conference on
  Computational Intelligence and Games}, 2015.

\bibitem{StreamLevels}
L.~N. Ferreira, ``Streamlevels: Using visualization to generate platform
  levels,'' \emph{ACM Computers in Entertainment}, 2015.

\bibitem{Jump_Thesis}
M.~Fasterholdt, ``You say jump, i say how high?'' Master's thesis, IT
  University of Copenhagen, 2015.

\bibitem{aramini:2017:thesis}
A.~U. Aramini, ``An ai assisted framework for the design of 2d platformers,''
  Master's thesis, Politecnico di Milano, Milano --- Italy, 2017.

\end{thebibliography}
\end{document}